# The thermodynamic cost of fast thought

Alexandre de Castro*

Laboratory for Computational Mathematics - LabMaC
Embrapa Agriculture Informatics - CNPTIA
Brazilian Agricultural Research Corporation - Embrapa
Av. André Tosello, 209, C.P. 6041, Campinas, São Paulo, 13083-886, Brazil

**Abstract**
After more than sixty years, Shannon's research continues to raise fundamental questions, such as the one formulated by R. Luce, which is still unanswered: "Why is information theory not very applicable to psychological problems, despite apparent similarities of concepts?" On this topic, S. Pinker, one of the foremost defenders of the widespread computational theory of mind, has argued that thought is simply a type of computation, and that the gap between human cognition and computational models may be illusory. In this context, in his latest book, titled *Thinking Fast and Slow*, D. Kahneman provides further theoretical interpretation by differentiating the two assumed systems of the cognitive functioning of the human mind. He calls them intuition (system 1) determined to be an associative (automatic, fast and perceptual) machine, and reasoning (system 2) required to be voluntary and to operate logical-deductively. In this paper, we propose a mathematical approach inspired by Ausubel's meaningful learning theory for investigating, from the constructivist perspective, information processing in the working memory of cognizers. Specifically, a thought experiment is performed utilizing the mind of a dual-natured creature known as Maxwell's demon: a tiny "man-machine" solely equipped with the characteristics of system 1, which prevents it from reasoning. The calculation presented here shows that the Ausubelian learning schema, when inserted into the creature's memory, leads to a Shannon-Hartley-like model that, in turn, converges exactly to the fundamental thermodynamic principle of computation, known as the Landauer limit. This result indicates that when the system 2 is shut down, both an intelligent being, as well as a binary machine, incur the same minimum energy cost per unit of information (knowledge) processed (acquired), which mathematically shows the computational attribute of the system 1, as Kahneman theorized. This finding links information theory to human psychological features and opens the possibility to experimentally test the computational theory of mind by means of Landauer's energy cost, which can pave a way toward the conception of a multi-bit reasoning machine.
**Keywords:** Computationalism; Ausubel's learning theory; Cognitive structure; Landauer limit; Maxwell's demon; Shannon-Hartley model.
*E-mail: alexandre.castro@embrapa.br

# Introduction

Pinker (2005)'s evolutionary arguments have suggested that thought can be understood as a form of computation. This philosophical perspective is known as the computational theory of mind and claims that an intelligent being works exactly as a computer works when it interprets an outside stimulus. However, such authors as Luce (2003) have asserted that the gap between psychology and information theory remains open, particularly when applying the concept of channel capacity to the context of the brain. From a viewpoint aligned with the notion of information in statistical theory (Laming, 2010) has presented the hypothesis that the human mind is a purely physical communication system. For Laming, the efficiency of brain performance directly depends on the connection between stimuli and channel characteristics.

From this same physicalist idea, Ladyman (2009) made a challenging statement asserting that if the brain implements computations as the computational theory of mind holds, the human mind necessarily must obey computational principles, such as the Landauer limit, that is a minimum cost energy per unit of information erased (processed) in a memory device (Landauer, 1961; Bennett, 1982; Bennett, 1987; Bennett, 2003; Maroney, 2009; Lambson et al. 2011).

In this context, several studies have (to date) reported an inability to find the "missing link" between psychology and information theory (Luce, 2003; Lenski, 2010). These studies have addressed the relationship between the presuppositions of intelligence and computation from a thermodynamic perspective only, but here we present an alternative path that starts from a psychological perspective and arrives at Landauer's thermodynamic limit, showing that this computational principle is the common frontier to both the cognitive and the machine worlds.

# Theoretical background

We have interpreted Ausubel's meaningful learning theory (Ausubel, 1963; Novak, 2002; Novak and Gowin, 2002; Seel, 2003; Smith, 2006; Ausubel, 2010; Moon et al.2011) as supporting a thought experiment (Buzzoni, 2008) regarding changes in mental state when information is meaningly assimilated by a cognitive system. Recent brain studies also lend support to the fundamental idea in Ausubel's constructivist theory that knowledge stored during meaningful learning is fundamentally organized differently than knowledge learned by memorizing process using routine or repetition (Novak, 2011).

According to the Ausubelian view, information becomes meaningful to an individual when it is "anchored" to relevant aspects (ideas or concepts) of the pre-existing cognitive structure, which are known as the subsumers.

Therese subsumers are constructs that can be thought of as idiosyncratic receptors that are equivalent to binding sites responsible for the specific recognition of new information. In addition to specific recognition, lock-key interactions between units of information and other less-related subsumers can also occur simultaneously, leading to mismatching. However, in accordance with Novak (2010), learning is more efficient when the initial external input is matched to the specific shape of the learner's framework. Therefore, an effective subsumption (assimilation) process occurs when units of information are associated in a complementary way to subsumers of the pre-existing structure, at which point the new information is internalized in the individual's framework. Both the information and subsumers are modified in an adaptive process, and the result of this subsumption is the learning of new concepts. The quantitative feature this learning is a new framework consisting of subsumers that are more differentiated with increased affinity to external stimuli. Ausubel describes this entire adaptive process as "progressive differentiation".

Based on the Ausubelian learning, the cognitive structure can be treated as a N-subsumers function, $S = S(s_1, s_2, ..., s_N)$. Changes to this structure occur through an "interactional products" (Ausubel, 1963; Seel, 2003; Ausubel, 2010) between external inputs $(I_1, I_2, ..., I_N)$ and the subsumers $(s_1, s_2, ..., s_N)$. Consequently, the number of units of information learned in a determined time interval is proportional to the sum of all the contributions of the matchings, $s_i I_j$, as well as the mismatchings. Thus, the units of information $I$ and the binding sites of cognitive structure $S = S(s_1, s_2, ..., s_N)$ should be complementary to some degree. In this work, this associative condition is represented by a similarity metric, $D_{s_i I_j}$.

In addition to the restructuring caused by external stimuli, the individual's framework undergoes continuous self-restructuring. This process is dynamic, and complementary relationships between multiple concepts (subsumers) already established in the individual's framework can also occur (Smith, 2006; Moon et al.2011). Certain elements in the framework can be perceived to be related to themselves. As a result, such relationships can also lead to structural self-reorganization. Ausubel describes this combination of previously existing elements in the cognitive structure, or self-reorganization provoked by self-interaction between subsumers, as an "integrative reconciliation" process. This process can be represented by a complementary function, $f(s,\bar{s})$, in which the subsumers of the $\bar{s}$ set have shapes that are complementary to those of the subsumers of the $s$ set. The meaningful learning is complete when integrative reconciliation occurs, i.e., when general concepts and their respective subsumers are differentiated and integrated wich each other. (Novak, 1977).

**A model for meaningful learning**

Ausubel (2010) wrote that the most natural way to acquire knowledge is through progressive differentiation; however, lateral connections between concepts can exist without the need for a hierarchy, thereby suggesting an integrative reconciliation of ideas. Novak and Gowin (2002) have asserted that this kind of integrative connectivity is associated with original and creative cognitive constructions in human beings.

Progressive differentiation and integrative reconciliation are clearly governed by two types of human cognitive processes (Kahneman, 2002; Kahneman, 2011) labeled system 1 and system 2, respectively. This mental behavior is a so-called dual-process[1] (Kahneman, 2002; Kahneman, 2011) which distinguishes system 1 (which creates an automatic activation based on such properties as similarity, pattern recognition and perception from external stimuli) from system 2 (which creates a conscious activation making use of logical principles). Based on this dual functioning of the mind, a human's meaningful learning can be mathematically modeled using the law of mass action (Jones et al. 2010), which is widely known and is used to describe relationships between biological entities. The implementation of this law is founded on the assumption that biological agents interact mathematically with susceptible populations through a multiplication operation, similar to the "interactional product" concept proposed by Ausubel for the subsumption process (Moreira, 1993; Moreira, 2011; Konrad, 2007; Seel, 2003).

---

[1] Pozo (2008), quoting Ausubel's theory, among others, also identifies two processes of learning: an automatic (effortless), and a controlled (effortful) that occurs voluntarily. These two ways to acquire knowledge are intrinsically (and clearly) attached to dual process theory of mind.

Thus, in the approach presented in this report, units of information and subsumers are treated in a manner analogous to the treatment of biological agents and susceptible populations, respectively. Consequently, the temporal behavior of an individual's mental structure can be defined by a dynamical approach, as follows:

$$\frac{d}{dt}S(s_1, s_2, ..., s_N) = \sum_{i,j=1}^{N} D_{s_i I_j} s_i I_j + f(s, \bar{s}), \quad (1)$$

where the $I_j$ terms are external inputs per unit time, and $s_i = s_i(t)$ are subsumers belonging to the pre-existing framework. The $\sum_{i,j=1}^{N} D_{s_i I_j} s_i I_j$ term represents the process of progressive differentiation governed by system 1; $D_{s_i I_j}$ is a normalized matching metric ($0 \leq D_{s_i, I_j} \leq 1$) that measures the affinity between $s_i = s_i(t)$ and $I_j$. The $f(s, \bar{s})$ additive quantity is an exchange term that represents the sum of the contributions of the self-interactions between subsumers, representing the process of integrative reconciliation governed by system 2, and $\frac{dS}{dt}$ is the modification rate of the cognitive structure per unit time as a result of the subsumption (assimilation) process. According to our approach of the law of mass action, this modification rate[2] replaces the Ausubelian interactional product[3], $sI$, in the final stage of the subsumption process.

Here, we consider that the binding sites of cognitive structure (subsumers) can be modeled by binary strings of length $\ell$ (Burgin, 2010; Seising, 2010; Hosoya et al. 2011). So, a subsumer can always recognize an external stimulus when there is complementarity between its strings. The matching (association) of information and subsumers is determined by considering the number of complementary shapes. For instance, if a subsumer in a 2-D space ($\ell = 2$) is represented by the string (0 1), and an input is represented by the string (1 0), a cognitive change takes place; however, the matching between strings does not need to be perfect (some mismatches are allowed). These differences between strings reflect the degree of affinity between them and determine the quality of the cognitive change[4], $\frac{dS}{dt}$.

---

[2] It is important to highlight that our contextual motivation for using a rate of modification of the cognitive structure (a metric over time that is directly proportional to the product of the information and the subsumers) was essentially inspired by Piaget (2001), because, according to Wadsworth (2003), the Piagetian assimilation process accounts for the growth of the intellectual structure that is provoked by, in his words, "a quantitative change". For his part, Novak (2010) claims that Piaget's assimilation process and Ausubel's subsumption process are elementally similar. In passing, Ausubel himself recognized a general similarity between Piaget's formulation of the assimilation process and his own assimilation theory (Ausubel et al., 1978; Nielson, 1980), which led us to infer that the Ausubelian subsumption can be quantitatively treated.

[3] Note that the Ausubelian interactional product, $sI$, which features the core of the learning schema by assimilation, is an associative (adaptive) representational procedure. Goldammer and R. Kaehr (1989) argued that adaptive learning can be represented by a Hebbian algorithm that simulates the linkage between the domain and the internal structure. On the other hand, Gerstner and Kistler (2002) claimed that Hebbian learning is a bilinear-type operation that represents a mathematical abstraction of coincidence detection mechanisms. So, taking into account that the internal structure is represented in our thought experiment by a space of subsumers, and that a domain space is just the boundary condition of this structure (i.e. the informational ambience), Ausubel's interactional product, consequently, takes the form of a bilinear isomorphism that combines elements of these two spaces, $s$ and $I$ (Konrad, 2007).

**Basal subsumption process**

Equation (1) is a differential model of several variables, representing a complex network of manifold connections between units of information and subsumers, and between subsumers and other subsumers; considering Ausubel's premises, this differential equation can be seen as a first-approximation hypothesis, an occam's razor-like model for human meaningful learning.

In this paper, equation (1) is studied in the context of a thought experiment for the simplest manifestation of intelligence, i.e., a clever being with only one subsumer that is able to process only a single unit of information per unit time. A mind with such characteristics is similar to the memory of the well-known Maxwell's demon[5] within the one-molecule engine described by Szilárd. This imaginary creature can extract information concerning the position of a single gas particle in an adiabatic box, and this position is the sole information that it can (learn) process. From a thermodynamic perspective, the mind of Maxwell's demon has been thoroughly investigated. Thermodynamically, Maxwell's creature is only a heat machine; it is a computing apparatus that can process information into its phase space, which constitutes its inanimate memory. However, from a constructivist perspective, Maxwell's demon in our thought experiment can also simulate a tiny intelligent being that is able to acquire knowledge.

---

[4] It is worth remembering that Ausubel's theory holds that new information is linked to relevant pre-existing aspects of cognitive structure with both the newly acquired information and the pre-existing structure being modified in the process — based on this conceptual model, Ausubelian theory describes the cognitive process of subsumption with its underlying principles of progressive differentiation and integrative reconciliation (Nielson, 1980). But, for this cognitive process occurs, the linkage between new information and pre-existing structure must be made by means of an Ausubelian interactional product, an Ausubelian matching.

[5] Maxwell's demon is a "clever" creature designed by James C. Maxwell with the extra-natural power to reduce the entropy of an isolated physical system, so violating the Second Law of thermodynamics. Maxwell's demon was "exorcised" by Szilárd in 1929 (Szilárd, 1925; Zurek, 1990).

According to the constructivist Ausubelian, a being that is able to learn must have a pre-existing mental structure that contains pre-designed concepts. Thus, using a one-subsumer approach for equation (1), Maxwell's demon mind should accomplish a basal subsumption of information, as follows:

$$\frac{dS}{dt} = D_{s,I} sI \,, \tag{2}$$

where the pre-existing structure of this creature is, *per se*, constituted by its single subsumer $S(s_1) \equiv s_1 = s$, and $I_1 = I$ is the unit of information matched by the demon's mind that is available in Szilárd's one-molecule engine. This unit of information, conditionally required by the law of mass action (Jones et al. 2010), is an input per unit of time. The condition $f(s, \bar{s}) = 0$ is also required because Maxwell's demon is assumed to be based on only one subsumer. Inasmuch as its framework is unable to relate multiple (or at least two) subsumers among themselves, this creature is non-reasoning. This demon can be thought of equivalently as a non-deductive being, which is governed only by cognitive system 1.

This way, one-subsumer equation (2) captures the essence of the Ausubelian principle of learning by progressive differentiation when a new unit of information, $I$, is linked to relevant pre-existing aspects of an individual's mental structure, $S$, and the pre-existing structure is modified in the process. In a Szilárd scenario (Maroney, 2009), the values of the variables that describe this subsumptive (perceptual) learning in terms of changes in the mental state of Maxwell's demon can be given by Brookes' procedural representation (Todd, 1999; Burgin, 2010). In his seminal "Fundamental Equation of Information Science," Brookes (1980) used a concept pointedly based on process of Ausubelian progressive differentiation. This representation, which can also be thought how a state transition, is defined by $K(S) + I \rightarrow K(S + \Delta S)$, wherein the state of mind $K(S)$ changes to another state $K(S + \Delta S)$, due to the input of one unit of information, $I$, and $\Delta S$ is an indicator of the effect this modification on the cognitive framework.

Given these formulations for a time-independent input, and accounting the limits of the integration provided by Brookes' representation, equation (2)

is solved as follows[6]:

$$\int_{\Delta t} I\,dt = \frac{1}{D_{s,I}} \int_{K(S)}^{K(S+\Delta S)} \frac{ds}{s}, \qquad (3)$$

where $K(S)$ and $K(S+\Delta S)$ are the a priori and a posteriori states[7] of Maxwell's creature, respectively[8] (Cole, 1997).

**Perceptual system as a purely physical communication channel**

Considering the entire subsumption process of information as contained within the perceptual system, the working memory capacity of this creature can be determined by $\Delta K = K(S+\Delta S) - K(S)$; consequently, equation (3) can be rewritten as follows[9]:

$$I = \frac{1}{D_{s,I}} \frac{1}{\Delta t} \ln\left(1 + \frac{\Delta K}{K(S)}\right) \qquad (4)$$

For the matching range ($0 \leq D_{s,I} \leq 1$), equation (4) can be recast as the following inequality:

$$I \geq \frac{1}{\Delta t} \ln\left(1 + \frac{\Delta K}{K(S)}\right) \qquad (5)$$

This logarithmic expression (5) is a Shannon-Hartley-like quantity (Burgin, 2010; Seising, 2010) with cognitive characteristics representing the processing capacity given in units of information per unit time, where a bandwidth in equation (4) can be identified as a function of both the subsumption time ($\Delta t$) and the noise/noiseless matching ($0 \leq D_{s,I} \leq 1$) between the external input and the mental structure of Maxwell's demon. In equation (5), $\Delta K$ represents the information storage, which has at least two distinguishable states of knowledge after the effect of an input (the left- or right-side position of the particle inside the bi-partitioned Szilárd's box), and $K(S)$ is a channel characteristic that represents a pre-existing memory state. This calculation, which is based on a constructivist treatment of Szilárd's one-molecule engine, supports the recent computationalist argument reviewed by Laming (2010), in which the main premise is to consider the human mind as a purely physical communication system whose performance depends on the match between external stimulus and channel characteristics[10].

---

[6] Here, $D_{s,I}$ appears outside the integral, because it is not a direct function of $s$ and $I$, but depends only on measure of the distance between a given information, $I$, and a reference subsume, $s$. This metric space is a number, a normalized length equal to 0 when $s$ and $I$ are maximally similar, and equal to 1 when $s$ and $I$ are maximally dissimilar (Li *et al.*, 2004).

[7] Brookes' equation represents the procedure for the retention of information described by Ausubel's theory of meaningful learning, when information and the subsumer can still be dissociated. This notation, considering the plus sign (+) in Brookes' equation, assumes the form of a superposition that indicates a cognitive change provoked by an external stimulus, leading the one-subsumer cognitive structure from state $S$ to state $S+\Delta S$. The consequence of this superposition concept is the immediate characterization of $K(S)$ as an eigenvalue associated with eigenstate $S$, and $K(S+\Delta S)$ as an eigenvalue associated with eigenstate $S+\Delta S$, which gives us the integration limits used in equation(3). In this context, it would be preferable to define the process of cognitive change as Brookes' superposition (or Brookes' state transition), rather than Brookes' equation.

[8] Cole (2011) claims that the use of information in information science is firmly linked to the information itself, which Brookes (1980) defined in his fundamental equation of information science as that which modifies a user's knowledge structure. In an earlier paper, Cole (1997) advocated a quantitative treatment for information based on Brookes' viewpoint. Bawden (2011) also suggests a possible quantitative treatment of Brookes' equation.

[9] For convenience, $\ln\left(\frac{K(S+\Delta S)}{K(S)}\right)$ was rewritten as $\ln\left(1+\frac{\Delta K}{K(S)}\right)$, because we not know the $\frac{K(S+\Delta S)}{K(S)}$ ratio, but the $\frac{\Delta K}{K(S)}$ term can be determined, as will be shown below.

**Information-to-energy conversion**

Taking into account recent works on information-to-energy conversion, in an isothermal heat bath as such Szilárd's one-molecule engine, information can be converted to energy up to $k_BTI$ (Sagawa&Ueda, 2010; Toyabe, *et al*., 2010). Thus, equation (5) can be subjected to a thermal treatment by scaling factor of $k_BT$, so that Maxwell's demon reaches the maximum efficiency[11]:

$$k_BTI \geq \frac{k_BT}{\Delta t} \ln\left(1 + \frac{\Delta K}{K(S)}\right), \quad (6)$$

where $k_B$ is the Boltzmann constant and $T$ is the absolute temperature of the microcanonical ensemble.

Denoting the amount of information per unit time, $I$, by *info*, equation (6) becomes

$$\left(\frac{info}{\Delta t}\right)k_BT \geq \frac{k_BT}{\Delta t} \ln\left(1 + \frac{\Delta K}{K(S)}\right). \quad (7)$$

It is known that in a thermally isolated one-molecule system, such as Szilárd's apparatus, $k_BT$ is the thermodynamic work to be accomplished for each step performed. More fundamentally, $k_BT$ is the Helmholtz free energy, a minimum amount of heat required per unit of information to increase the entropy of Szilárd's apparatus by one unit of information (Levitin, 1998; Plenio&Vitelli, 2001; Toyabe, et al., 2010; Sagawa&Ueda, 2008).

So, equation (7) can be rewritten as follows:

$$\left(\frac{info}{\Delta t}\right)\Delta E \geq \frac{k_BT}{\Delta t} \ln\left(1 + \frac{\Delta K}{K(S)}\right), \quad (8)$$

---

[10] Brookes (1981) also suggested a logarithmic expression to solve his equation. From a perspectivist approach, he outlined a skeleton of an equation to represent the carrying of information into the human mind on the same basis as Hartley's law (Castro, 2013).

[11] Toyabe, et al.(2010) have achieved the $k_BTI \geq k_BT \ln(\ )$ relationship for a quasistatic information heat engine such as the Szilárd engine – in their notation, $I$ is equivalent to $I\Delta t$, in our calculation.

Rearranging the terms[12],

$$\Delta E \geq k_B T \ln\left(1 + \frac{\Delta K}{K(S)}\right) \text{ per info}. \quad (9)$$

Still, considering the working memory capacity, $\Delta K = K(S + \Delta S) - K(S)$, and that in Brookes' schema the state of mind $K(S)$ changes to another state $K(S + \Delta S)$ due to the input of one unit of information, the ratio between demon's mental states can be given by $\frac{K(S)}{K(S+\Delta S)} \geq e^{-\frac{\Delta E}{k_B T}}$, where $e^{-\frac{\Delta E}{k_B T}}$ is the Boltzmann weighting factor (Kittel, 1980). From the Boltzmann factor it is possible to infer a quantum behavior of intuition-system, as will be emphasized in footnote (16).

**Forgetting of information**

In Szilárd's hypothetical world, the position of the particle inside the bi-partitioned box is all that Maxwell's demon can learn. This limitation restricts the working memory capacity of the creature such that it can only acquire a single unit of information at a time. However, considering Maxwell's demon as a clever being, the Ausubelian subsumption process of information can be employed to restore its mind, but two essential stages referred to as "retention" and "obliterative" must be taken into account[13].

These two cognitive stages are drawn by Brookes' equation of Ausubelian subsumption process of information[14], $K(S) + I \quad K(S)I \to K(S + \Delta S)$, where $I$ is retained (anchored) in $K(S)$ by means of an interactional product that yields an unstable quantity, $K(S)I$. Over time, the $K(S)I$ quantity is no longer dissociable, and the subsumer is modified to acquire additional subsumed meanings. As a result, the obliterative phase arises and a "forgetting" occurs, in which the information is internalized into the new stable structure, $K(S)I \to K(S + \Delta S)$ [15]. At the end of the obliteration process, $I$ is absorbed into the mental space, emptying the working memory again in order to allow the assimilation of new information by Maxwell's demon – in Ausubel's words himself, $K(S)$ and $I$ are "reduced to the least common denominator" (Ausubel,1963; Ausubel, 2010; Brown, 2007).

---

[12] Whereas $\Delta E = \frac{heat}{info}$, such as shown by Levitin (1998), for that $k_B T$ is a minimum amount of heat per unit of information, the $\frac{\ln(\cdot)}{info} = 1$ condition must then be satisfied.

[13] In Ausubel's theory, meaningfull learning occurs when the individual creates a connection between information, $I$, and subsumer, $S$, resulting in an "interactional product" $sI$ (Moreira, 2011). The notion of subsumption of information in the Ausubelian meaningfull learning is that $s$ and $I$ remain dissociated from the product $sI$ during the first stage of assimilation process of information and, over time, $I$ becomes closely linked with $s$, being anchored in the ideational complex $sI$.

[14] According to Todd(1999), Brookes also assumed that the unit of knowledge is concept based, a notion derived from Ausubel's assimilation theory of cognitive learning.

[15] During the process of assimilation, the new meaningful gradually loses its identity as it becomes part of the modified anchoring structure. This process is termed obliterative

**Landauer limit from a cognitive squeeze**

The obliterative phase of subsumption is nothing more than the mapping of two mental states, $K(S)$ and $I$, into one, $K(S+\Delta S)$. Consequently, Ausubelian forgetting yields a cognitive squeeze[16], applying a relative change to demon's mind that is equal to $\frac{\Delta K}{K(S)}=-\frac{1}{2}$. Inserting this cognitive squeeze[17] into equation (9) and considering Shannon's binary digit as basic unit of information in computing, we obtain from the cognizer's mind the quantity known as the Landauer limit (Plenio&Vitelli, 2001):

$$\Delta E \geq -k_B T \ln(2) \, per \, bit, \qquad (10)$$

or ultimately, $\Delta E \geq -k_B T \, per \, nat$, where, $1 \, bit = \ln(2) \, nat$ (Levitin, 1998; Sagawa&Ueda, 2010).

In "Waiting for Landauer," Norton (2011) claimed "Landauer's principle asserts that there is an unavoidable cost in thermodynamic entropy creation when data is erased. It is usually derived from incorrect assumptions, most notably, that erasure must compress the phase space of a memory device ...". However, in this paper, we show that the Landauer limit given by equation

---

subsumption and is dependent on dissociability strength between $s$ and $I$. This gradual loss of separable identity ends with the meaning being forgoteen when the idea falls below the "threshold of availability" proposed by Ausubel. (Seel,2012).

[16] Brookes' equation shows that the two cognitive states, $K(S)$ and $I$, are compressed into one state, $K(S+\Delta S)$. Thus, the $\frac{\Delta K}{K(S)}$ term represents a relative cognitive change, a squeeze cognitive (a reduction) equal to - 50%. Consequently, the Boltzmann factor obtained from equation (9) is $e^{-\frac{\Delta E}{k_B T}} \leq 2$, since $\frac{K(S+\Delta S)}{K(S)}$ is equal to $1+\frac{\Delta K}{K(S)}$. This weighting factor evaluates the relative probability of a determined state occurring in a multi-state system, i.e., it is a "non-normalized probability" that needs be $\gg 1$ for the system to be described for a non-quantum statistics; otherwise, the system exhibits quantum behavior (Carter, 2001). So, the Boltzmann factor ≤ 2 calculated here indicates that the subsumption process requires a quantum statistics to describe the cognitive learning. In other words, our calculation suggests that the final stage of the subsumption process of information – within the Kahneman's intuition-system – produces a quantum quantity, as Aerts(2011) has advocated. Toyabe, et al.(2010) and Sagawa&Ueda (2008) also have achieved this same value for the Boltzmann factor in an feedback control system, however, they raised this outcome by means of physics; here, we were able to achieve this outcome by means of psychology.

[17] As an external stimulus is subsumed into an individual's knowledge structure – which, from Bennett's thermodynamics viewpoint, corresponds to the deletion of information from memory devices – the relative cognitive change $\left(\frac{\Delta K}{K(S)}\right)$ assumes negative values, indicating a cognitive squeeze.

(10) appears in a cognoscitive world after a "cognitive squeeze" that is required to complete the subsumption process. This outcome is similar to that found from a thermodynamic perspective when Bennett's (compression) erasing is processed into the mind (computer memory) of Maxwell's demon (Bennett, 1982; Bennett, 1987; Bennett, 2003).

In Bennett's words, "...from its beginning, the history of the Maxwell's demon problem has involved discussions of the role of the demon's intelligence, and indeed, of how and whether one ought to characterize an intelligent being ..." (Bennett, 2003). In this study, our calculation instigates two interpretations: (a) when Maxwell's demon is interpreted as a classical machine, the demon is able to process information in agreement with Ausubel's meaningful learning theory, indicating the possibility that a machine can learn how a human learns; (b) when Maxwell's demon is interpreted as an intelligent being equipped solely as a fast system of type 1, it encounters the fundamental thermodynamic principle of computation (the Landauer limit), as an immediate consequence of the Ausubelian obliterative process; which shows the computational feature of a thinking system of type 1.

As soon, if computational presuppositions are essential conditions for the plausibility of the computational theory of mind, this dualist viewpoint now provides at least one reason for the plausibility: Landauer's thermodynamic limit is shared between cognizers and machines.

**Landauer efficiency**

In a recent work, Khemlani and Johnson-Laird (2012), citing Kahneman's last work (2011), wrote that "Cognitive psychologists similarly distinguish between system 1, which makes rapid automatic inferences based on heuristics, and system 2, which makes slower conscious deliberations based on systematic and perhaps normative principles. What is missing from these theories is an algorithmic account of the processes on which system 1 relies". In the present work, we have mathematically shown that system 1 relies on Landauer's thermodynamic principle, which indicates that it works exactly as a programmable machine designed to carry out perceptual operations sequentially and automatically. This result, which is based on cognitive premises, strongly suggests that the system 1 addressed in Kahman's theory is actually similar to an associative machine, quantitatively evidencing the computational theory of mind in terms of the thinking system 1. In addition, if system 1 is considered to be a powerful computer operating at maximum Landauer efficiency – i.e., at a minimum energy cost equal to $k_B T \ln(2)$ – that works at an average brain temperature, the number of perceptual operations per second that it could perform is on the order of $10^{23}$ ($1/k_B$), depending on the idiosyncratic power of the brain. This rate of operation is quite fast, as Kahneman has theorized, and it is also inexpensive!

**Computational theory of mind: an experimental outlook**

The psycho-philosophical hypothesis that thought is a kind of computation, if supported by experimental evidence, would imply that a substantial amount of disparate research fields, from cognitive psychology to information theory, could be unified, and that a new formalism could be introduced for broader consideration. Taking into account this scenario, the calculation presented here offers an innovative standpoint — an energy perspective — regarding the foundations of human thought, and it can circumstantially provide an observable mode to experimentally ascertain the controversial computational theory of mind that has so far been lacking in quantitative arguments, in particular regarding whether or not a Turing-like algorithm is the operational *substratum* underlying the perceptual system (Davenport, 2012; Swiatczak, 2011; Pinker, 2005; Fodor, 2000). It is worth remembering that, although the Turing machine constitutes the basis of the debate on computationalism (Fresco, 2012), such a stored-program device is considered only an abstraction of a computing apparatus, a nonphysical framing, i.e. Turing's machine cannot be subject to experimental verification.

However, quite recently, Frasca (2012), in a preprint entitled, "Probabilistic Turing Machine and Landauer Limit", argues that Landauer's energy boundary can be obtained from a Turing machine[18]. This Frasca's inference is in consonance with our conception of the *vinculum* between the Landauer limit and the computational theory of the mind[19] and, as a result of that relationship, the theoretical essay presented in this paper becomes a possible (and challenging) alternative to experimentally test the computationalist hypothesis of the associative cognitive process, since the Landauer limit can be empirically checked[20].

---

[18] This singular consequence reinforces the idea that Landauer limit is an intrinsic feature of algorithmic rules, rather than a quantity associated to physical part of computer.

[19] For certain, we are not asserting that the human mind, as a whole, operates like an algorithmic machine, but, when system 2 was left out of our calculation for the reason explained above, the outcome pointed precisely to Landauer's principle of thermodynamics, which – considering Frasca's work – mathematically indicates that the adaptive (associative) learning process (notably Ausubel's progressive differentiation) works in the same way as a step-by-step computational procedure.

**Concluding remarks**

In this paper, the meaningful learning process (i.e., without considering Ausubel's rote learning) was divided into two parts in equation (1). The first part refers to an association of external information with binding sites (subsumers) of an individual's cognitive structure, which is an involuntary process governed by system 1. The second refers to interaction between subsumers, representing a self-interaction, the capacity to deduce, and to voluntarily use the logical process; such a capacity is governed by system 2. An example of a system 1 event in the context of human learning is the presentation of a word that "automatically" elicits other concepts (subsumers) that are associated with it in cognitive structure. In a perceptual process, the operating characteristics of system 1 involve the use of previous knowledge to gather and interpret the external stimuli registered by the senses. Within system 1, stimuli are matched (associated) to a set of templates (patterns) stored in memory, and when a match is located, the new knowledge is learned.

From this conceptualization, we proposed the equation (2), solving it from the scenario of Szilárd's one-molecule engine (a variant of Maxwell's demon thought experiment) to support a single-subsumer-like approach, thereby causing the second part of equation (1) to vanish – namely, disregarding the part denotes which Fodor (2000) calls non-modular system. This approach supports the notion of a thinking creature, although non-reasoning, that is able to use only the perceptual system – or modular system in Fodor (2000)'s view – to process a unit of information at time. The calculation presented in this report shows that the obliterative phase of human meaningful learning proposed by Ausubel works in the same way as the process of thermodynamic compression proposed by Landauer (1961) and Bennett (1982; 1987; 2003), subject to the same energetic cost as a machine to restore its binary memory. In other words, from the Maxwell-Szilárd apparatus, we have shown that each bit processed by a cognizer needs one Ausubelian obliteration, and a minimum amount of energy per bit is required in this process. This minimum energy cost (the Landauer limit) returned to the environment of Maxwell-Szilárd's demon after a thermodynamic compression is a thermodynamic principle associated with the erasure (discarding) of information by binary machines, specifically, a one-bit logical operation that restores 0 and 1 to 0, which corresponds to the obliterative process in cognizers. This compression process is necessary for a binary device to reset its memory (Landauer, 1961; Lambson et al. 2011; Bérut et al. 2012), and Bennett (1982; 1987; 2003) has explained this process entirely for computers from the thermodynamic viewpoint. The novel finding presented in this report is that we have explained the same process from the cognitive viewpoint, which gives rise to a new perspective in terms of artificial intelligence. Ahead of this finding, the great challenge seems to be solving equation (1) when it also includes the second term that represents the system 2, what is the human being itself. Thereby, we strongly believe that the study of a full solution for this equation can open a new protocol, which will allow the transfer of the principles of human reasoning to computers.

---

[20] Recently, a group of researchers was able to experimentally show Landauer's principle in a two-state bit memory (Bérut et al., 2012). It is therefore plausible to presume that Landauer's principle in human mind can also be subjected to experimentation. Landauer's limit is an energy boundary, in the form of heat dissipation – a physical quantity, which can provide a robust evidence of an algorithmic-like activity within the "gears" of a type 1 cognitive system. In addition, Landauer's boundary can also quantitatively underpin latest Kahneman's qualitative presuppositions for a intuition-system (Morewedge&Kahneman, 2010).

**Acknowledgments**. This work was supported by the Centro Nacional de Pesquisa Tecnológica em Informática para a Agricultura (Empresa Brasileira de Pesquisa Agropecuária - Embrapa).